\documentclass{article}

\usepackage[preprint]{neurips_2026}

\usepackage[utf8]{inputenc}
\usepackage[T1]{fontenc}
\usepackage{hyperref}
\hypersetup{hidelinks}
\usepackage{url}
\usepackage{booktabs}
\usepackage{multirow}
\usepackage{amsfonts}
\usepackage{amsmath}
\usepackage{amssymb}
\usepackage{microtype}
\usepackage{xcolor}
\usepackage{float}
\usepackage{pgfplots}
\pgfplotsset{compat=1.18}
\usetikzlibrary{positioning, arrows.meta, fit, backgrounds, decorations.pathreplacing}

\newcommand{\dR}{\Delta R}
\newcommand{\sa}{\sigma_a}
\newcommand{\sep}{\sigma_{\mathrm{ep}}}
\newif\iffulldisclosures
\fulldisclosuresfalse

\title{Measuring the Value of World-Model Updates\\
A Counterfactual Utility Protocol for Continual Adaptation}

\author{%
  Anqi Peter Li\thanks{Correspondence: \href{mailto:peterli@substrate-labs.org}{\texttt{peterli@substrate-labs.org}}}\\
  Substrate Labs
  \And
  Kaden Kim\\
  UC Berkeley
}

\begin{document}

\maketitle

\begin{abstract}\noindent
Continual world models must decide whether new data justify changing the model. Fixed replay schedules
and prediction-error triggers specify when to update, but neither reveals the value of an individual
update: one deployment run cannot show how the same model would have performed at that moment had
it held its parameters. We introduce the \emph{fork ledger}, which branches a deployment stream at
pre-registered decision points into matched \textsc{update} and \textsc{hold} continuations under
common random numbers. It evaluates both continuations on the same episodes and records
$\dR = R_{\textsc{update}} - R_{\textsc{hold}}$. Always applying one fixed update mechanism lowers
return on all three simulated control tasks: CartPole ($-144.0$; checkpoint-bootstrap 95\% CI
$[-185.4,-116.1]$, against a converged return near $650$), Walker ($-82.8$; $[-101.1,-61.7]$) and
Cheetah ($-18.6$; $[-29.0,-6.6]$). Divergence is an outcome of applying the update, so the
estimand counts every attempted fork; restricted to the $693$ of $720$ that did not collapse,
CartPole and Walker are unchanged in sign ($-113.4$ and $-82.1$) and Cheetah becomes unresolved
($-3.9$; $[-17.5,+13.0]$). The task is the unit of inference: each contributes $240$ attempted
forks over five pretrained checkpoints crossed with two drift directions.
The ledger makes counterfactual utility observable for a fixed mechanism, allowing triggers to be
judged by the updates they select rather than by surprise detection alone.
\end{abstract}

\section{Introduction}
\label{sec:intro}

Continual world models must decide when new data justify changing the model. DreamerV3 uses a fixed
replay ratio; deployment-time adaptation can finetune a world model online, while model-management
methods use model-shift, under- or overfitting, or change-point signals to decide when to update
~\citep{hafner2023dreamerv3,feng2023fowm,ji2022cmlo,dorka2023dutd,vovk2021retrain}. These lines of work specify conditions for updating; the fork
ledger asks whether an individual update improves control.

The missing quantity is counterfactual. Prediction error mixes reducible model mismatch with
irreducible transition uncertainty, and an update can improve recent loss while degrading the
representation used for control. At a decision point, the relevant comparison is therefore the return
after updating versus the return the same deployed model would have obtained had it held its
parameters. Existing model-management and retraining policies frame the decision around aggregate behavior,
detection outcomes, or forecasted model quality rather than this matched per-update control contrast
\citep{liebman2018stitch,regol2025retrain,steland2026online}.

The \emph{fork ledger} makes the comparison observable in simulation. At each pre-registered decision
point, it runs matched \textsc{update} and \textsc{hold} continuations under common random numbers
and records $\dR_k = R_{\textsc{update},k}-R_{\textsc{hold},k}$ (Section~\ref{sec:ledger}). In the
corrected sweep, always applying the one fixed update mechanism we study lowers return on
CartPole and Walker, while Cheetah depends on drift direction and is unresolved. The stream-seed replication preserves this pattern. We therefore make a measurement claim
and a task-bounded empirical claim, rather than propose a universal update rule.

\paragraph{Contributions.}
\begin{itemize}\setlength{\itemsep}{0pt}\setlength{\parskip}{0pt}
\item The fork ledger, a reusable matched counterfactual protocol that assigns a realized value to
an individual update.
\item Across three tasks, the independent unit of inference ($n_{\text{task}}=3$, plus a
legacy-schedule negative control), each with $240$ attempted forks over five pretrained
checkpoints and checkpoint-clustered intervals: one fixed rehearsal-free update mechanism is
reliably harmful on all three, and its size changes with task and drift.
\item A direct comparison of pre-fork trigger signals against the same labels. Residual and
deployment-return triggers, the only families scored as decision policies in the corrected arm,
are nearly tied on CartPole, where their ordering is largely a competence measurement, and neither
ranking transfers across tasks.
\end{itemize}

\section{Related Work}
\label{sec:related}

\textbf{Detecting and timing updates.} Novelty and out-of-distribution methods study when a learned
model is unreliable~\citep{zollicoffer2025novelty,nasvytis2024rethinking}. Event-triggered learning,
concept-drift management, and retraining policies study when to adapt, assessing detection quality
or aggregate downstream behavior
\citep{ji2022cmlo,dorka2023dutd,vovk2021retrain,liebman2018stitch,regol2025retrain,steland2026online}.
Off-policy evaluation can estimate the
counterfactual, but only under assumptions about the response mechanism
\citep{oberst2019counterfactual,buesing2019cfgps}; in simulation, both branches can instead be run.

\textbf{Residuals, adaptation, and update value.} Likelihood and prediction-error signals can conflate
model mismatch with data entropy or irreducible noise
\citep{nalisnick2019know,caterini2021entropic,zhang2021understanding,pathak2019disagreement,
mavorparker2022noisytv,jarrett2023hindsight}, and one-step likelihood need not track control
performance~\citep{lambert2020objective,predictability2026control}. Recent systems use residual,
consistency, or online-world-model signals for adaptation or evaluation
~\citep{domberg2026selfadapting,coco2026,liu2025continualwm}.
Closest is \citet{twinrollouts2026}, which couples noise across branches of a generative video
model but intervenes on the action stream, where abduction is exact because the factual branch is
self-generated; the fork ledger intervenes on the parameters, couples branches through simulator
state and three randomness streams, and scores realized control return rather than plasticity
loss or detection accuracy.

\section{Method}
\label{sec:method}

\subsection{Problem setting}
\label{sec:setting}

\paragraph{Deployment stream.} A world model $p_\theta$ is pretrained on a stationary
environment and then deployed on a stream in which the environment changes. Two stream properties
must be distinguished because a residual conflates them. \textbf{Volatility} is drift in a
physics parameter over time (gravity for \texttt{cartpole}, contact friction for \texttt{walker});
an update can, in principle, track it. \textbf{Stochasticity} is Gaussian noise of scale $\sa$ added
to the executed action; it raises transition uncertainty but is not an environmental change that an
update can remove. We match drift profiles on accumulated level and rate so that residual
detectability, rather than an unintended schedule difference, does not drive the comparison.

\paragraph{Control depends on the world model.} Control uses latent-space MPC with cross-entropy-method
(CEM) action-sequence optimization~\citep{rubinstein1999cem}, with a terminal value bootstrap in the
style of TD-MPC2~\citep{hansen2024tdmpc2}. This choice matters: an amortised actor trained on the
old latent space would confound a world-model update with actor mismatch, whereas MPC consults the
current model at decision time, so an update changes behavior without actor retraining.

\paragraph{The update mechanism is held fixed.} Every paired continuation applies $N$ gradient steps
to the most recent $W$ transitions. We study \emph{trigger times}, not competing update mechanisms;
the labels are consequently relative to this response. A $100$-cell mechanism sensitivity changes the
verdict only in the one task--direction combination where updating gains (Appendix~\ref{app:mech}).
The mechanism rehearses nothing: the deployment buffer is allocated fresh when the stream starts and
the pretrained checkpoint stores model and optimiser state only, so the $W$-transition window holds
post-deployment data exclusively. The $\dR$ labels therefore price no-rehearsal fine-tuning of a
converged model, the regime in which catastrophic forgetting and unwanted drift are expected,
not update value in general.

\subsection{The fork ledger}
\label{sec:ledger}

The protocol has three stages (Figure~\ref{fig:ledger}).

\paragraph{1. Reference stream.} A pretrained world model is deployed for $T=6{,}000$ control
steps ($K=24$ decision points spaced $250$ steps apart, at an action repeat of $4$) under one
cell: one pretrained checkpoint and one drift direction, at $\sa=0$ throughout the corrected arm. We log the one-step
prediction error stream, the raw (unclipped) KL, the drift level $g(t)$ and rate $\Delta g(t)$,
and the injected noise level. Freezing makes the same stream reusable for every trigger: an
updated model's error stream would depend on the trigger that produced it.

\begingroup
\setlength{\intextsep}{4pt}
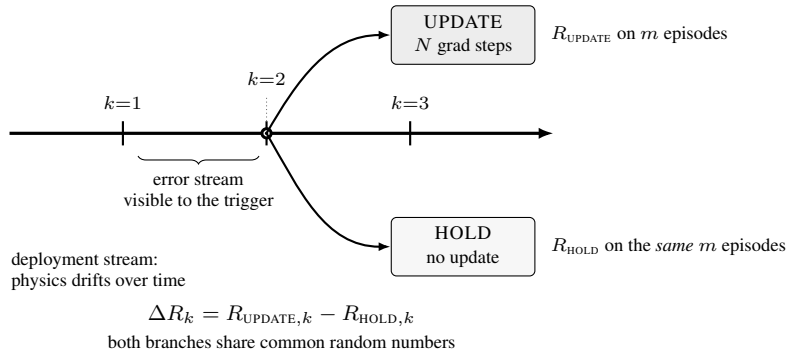
\begin{figure}[b]
\centering
\begin{tikzpicture}[
  font=\small,
  branch/.style={draw, rounded corners=2pt, minimum height=6.5mm, minimum width=19mm, align=center},
]
% ---- the deployment stream -------------------------------------------------
\draw[very thick, -{Latex[length=2mm]}] (0,0) -- (7.2,0);
\node[anchor=north west, font=\scriptsize, align=left] at (-0.1,-1.45)
  {deployment stream:\\physics drifts over time};

% decision points, labelled ABOVE the axis and clear of the fork curves
\foreach \x/\k in {1.5/1, 3.4/2, 5.3/3} {
  \draw[thick] (\x,-0.16) -- (\x,0.16);
}
\node[anchor=south, font=\scriptsize] at (1.5,0.20) {$k{=}1$};
\node[anchor=south, font=\scriptsize] at (5.3,0.20) {$k{=}3$};

% ---- the fork at k=2 -------------------------------------------------------
% The branch point is marked explicitly. Without a dot the curves appear to leave
% from somewhere between k=2 and k=3 and the reader cannot pin the fork down.
\node[anchor=south, font=\scriptsize] at (3.4,0.52) {$k{=}2$};
\draw[densely dotted, black!55] (3.4,0.16) -- (3.4,0.50);
\fill (3.4,0) circle (2.3pt);
\draw[white, line width=0.5pt] (3.4,0) circle (0.9pt);

\node[branch, fill=black!7] (upd) at (6.0, 1.30) {\textsc{update}\\[-1pt]{\scriptsize $N$ grad steps}};
\node[branch, fill=black!3] (hld) at (6.0, -1.50) {\textsc{hold}\\[-1pt]{\scriptsize no update}};
\draw[-{Latex[length=1.6mm]}, thick] (3.4,0) to[out=60,in=180] (upd.west);
\draw[-{Latex[length=1.6mm]}, thick] (3.4,0) to[out=-60,in=180] (hld.west);

\node[anchor=west, font=\scriptsize, align=left] at (upd.east) {$\;R_{\textsc{update}}$ on $m$ episodes};
\node[anchor=west, font=\scriptsize, align=left] at (hld.east) {$\;R_{\textsc{hold}}$ on the \emph{same} $m$ episodes};

% ---- pre-fork observation window ------------------------------------------
\draw[decorate, decoration={brace, amplitude=3pt, mirror}] (1.7,-0.30) -- (3.3,-0.30);
\node[anchor=north, font=\scriptsize, align=center] at (2.5,-0.42) {error stream\\visible to the trigger};

% The contrast is the figure's conclusion, so it sits at the bottom rather than
% beside the axis where it competed with the timeline for attention.
\node[anchor=north, align=center, font=\footnotesize] at (3.6,-2.15)
  {$\dR_k = R_{\textsc{update},k}-R_{\textsc{hold},k}$\\[1pt]
   {\scriptsize both branches share common random numbers}};
\end{tikzpicture}
\caption{\textbf{The fork ledger.} A deployment stream is checkpointed at pre-registered
decision points $k$ and branched into \textsc{update} and \textsc{hold} continuations, evaluated
on the \emph{same} $m$ episodes under common random numbers; $\dR_k$ is the realized value of
updating at $k$. A trigger observes only the pre-fork error stream (brace), so every candidate is
scored against the same labels without rerunning the environment.}
\label{fig:ledger}
\end{figure}
\endgroup

\paragraph{2. Paired labels.} At $K$ pre-registered decision points, we checkpoint and run two
branches. \textsc{update} applies $N$ gradient steps to the world model on the most recent
$W$ transitions, then evaluates. \textsc{hold} evaluates immediately. Both branches are
evaluated on the same $m$ episodes with identical initial states, injected action noise, and
planner sampling noise. The physics perturbation is \emph{pinned} at its
value at the moment of the fork: return is measured under the world as it stands at the decision
point, not one that keeps changing during evaluation. This choice has a cost. Pinning biases
the contrast along the volatility axis, under-crediting updates where drift is faster if control loss
is convex in staleness and \emph{over}-crediting them if it saturates.
Appendix~\ref{app:bias} quantifies both directions; which one obtains is empirical, not assumed,
and the forward-window arm that bounds it empirically was run only on CartPole.

\paragraph{3. Scoring.} We define $\dR$ as the causal effect of \emph{applying the fixed update
mechanism now} versus \emph{not applying it}: it prices the entire intervention---timing, dose $N$,
window $W$, and the induced model change together---and does not separate those components; it is
not a timing effect with the dose partialled out. This is deliberate: it matches the decision a deployed trigger
makes. Consequently, the base rate $\Pr(\dR>0)$ carries a dose component, so the precision endpoint
should be read against that base rate rather than against zero. A trigger is any function of
the pre-fork error stream: at a chosen rate it partitions decision points into triggered and
untriggered cases and is scored against the fixed $\dR$ labels, as pure post-processing. For the
corrected signal comparison, the deployment-return feature is the mean return from the segment
immediately before the fork, recorded before either branch is evaluated. We never use the post-fork
$R_{\textsc{hold}}$ column as a trigger feature. Common random numbers are standard
practice~\citep{glasserman1992crn}; three separate noise sources had to be pinned before repeated
runs agreed.

\subsection{Design corrections and pre-registered plan}
\label{sec:design}
\label{sec:preregistration}

\paragraph{Design and inference.} \emph{Corrected} cells run the committed protocol: a
$20{,}000$-step drift period, forks every $250$ control steps, and a logged pre-fork
deployment-return field. \emph{Legacy} cells predate one or both; they supply sensitivity only,
and those without the pre-fork field are quarantined from trigger ranking.
The pre-registered design makes five choices explicit. Its primary endpoint,
$\mathbb{E}[\dR\mid\text{triggered}]-\mathbb{E}[\dR]$ in units of the per-episode return standard
deviation $\sep$, is not reported here, a deviation; mean cell-level $\dR$, checkpoint-clustered
AUC and policy totals stand in its place. A trigger reads only the pre-fork
stream. AUC, whether a pre-fork scalar orders positive-$\dR$ forks above negative ones, is
computed per checkpoint over its two pre-registered direction streams combined, and aggregated by
checkpoint-cluster bootstrap rather than pooled over raw forks
\citep{cameron2008clusterboot}. All corrected cells run $\sa=0$ with $m=30$ matched evaluation
episodes per branch, the registered episode ramp not having been executed, so the stochasticity
axis is unexercised. With only five clusters the percentile bootstrap under-covers; Student-$t$
intervals on the same checkpoint means leave both negative verdicts intact
(Appendix~\ref{app:slope}), and no multiplicity adjustment is applied across the signal
comparisons, which we read descriptively. We compare against a
rate-matched random trigger, the $\dR$ oracle, and the registered drift oracle, which is
simulator-only and so is a ceiling rather than a deployable rule. The registered design specified causal running-quantile
thresholds; as a deviation from that registration, the corrected results instead fit the threshold
level leave-one-checkpoint-out on the pre-fork signal. The rate-matched control permutes within each held-out cell, preserving the
number of fires the reported policy made there; restricting it further to contiguous time blocks
leaves the verdict unchanged wherever the restriction leaves the null any freedom
(Section~\ref{sec:trigger})~\citep{anderson2003permutation}. The
registered choices and the deviations from them are recorded in
Appendices~\ref{app:prereg} and~\ref{app:dev}.

\paragraph{Environment and model.} The main study is proprioceptive \texttt{cartpole-swingup} from
the DeepMind Control Suite~\citep{tassa2018dmc} on MuJoCo~\citep{todorov2012mujoco}, with
\texttt{walker-walk} and \texttt{cheetah-run} replications and a legacy-schedule
\texttt{hopper-hop} negative control (Section~\ref{sec:cost}). The model is a compact DreamerV3-style recurrent state-space model
\citep{hafner2023dreamerv3}, trained from scratch on clean dynamics. The trainer reads back the
constructed-environment flags to verify no perturbation or action noise was active during
pretraining. The corrected grid and the registered trigger families are listed in
Appendices~\ref{app:prereg} and~\ref{app:dev}.

\section{Results}
\label{sec:results}

We report $\dR$ continuously rather than binarizing it into a precision label, which would attenuate
the effect of noisy fork outcomes; the primary protocol and robustness checks are in
Appendix~\ref{app:notes}.

\subsection{A fixed update rule loses return on all three tasks}
\label{sec:cost}

The ledger turns an update rule into a policy over forks, allowing policies to be compared in return
rather than detection rate (Figure~\ref{fig:policy}). In the corrected CartPole sweep the mean cell-level effect over all $240$
attempted forks is $-144.0$ $[-185.4,-116.1]$ against a converged return near $650$. Excluding diverged forks conditions on a
post-treatment variable, so the all-attempted mean is the estimand for update value and the
retained mean answers the narrower question of what a non-collapsing update is worth. Under the
divergence rule, which drops any fork whose update branch scores below $30\%$ of its cell's
frozen-branch reference, $228$ remain, updating helps on $58$ and hurts on $170$, the mean is
$-113.4$ $[-131.2,-90.1]$, and all ten cells are negative. The harm is largest where drift is smallest: mean $\dR$ in the lowest
drift quartile is $-208.6$ toward and $-253.4$ away, against $-8.4$ and $+53.0$ in the highest.
The two directions traverse the drift range in opposite time order, so this is not stream
position; the measured cost is not a failure to track drift.

The additional tasks bound the claim (Table~\ref{tab:multitask}). On Walker the two estimands agree, $-82.8$
$[-101.1,-61.7]$ over all attempted forks and $-82.1$ $[-100.2,-60.9]$ retained, with all ten
cells negative under both drift directions. Cheetah is where they part: all attempted gives $-18.6$
$[-29.0,-6.6]$, reliably negative, while the retained subset spans zero ($-3.9$ $[-17.5,+13.0]$)
with its sign set by the direction the world moved, $-26.8$ toward the pretraining dynamics and
$+19.0$ away. The rule biases $\dR$ toward zero, so the retained verdict is the conservative one.

A fourth task tests the instrument rather than the main claim. On \texttt{hopper-hop}, the
legacy-schedule control, the planner never leaves the floor, so an update has little to change. The ledger reports $+1.0$ $[-0.3,+2.4]$
against the registered bound $|\dR|<2.2$; the point estimate is inside the bound, although the upper
endpoint is not. One caveat scopes this control: Hopper ran under the legacy drift schedule
(period $2{,}000$ steps, saturating from the first fork in the \emph{away}
direction) rather than the corrected
$20{,}000$-step schedule, so its drift-direction split is not comparable to the rows above; within
that schedule, the control gives no evidence of a systematic negative shift.

The divergence rule ($12$ CartPole, $2$ Walker, $13$ Cheetah forks) is applied before any trigger
is scored, and it conditions every reported effect on the forks it retains
(Section~\ref{sec:discussion}). A separate legacy $46$-cell CartPole dose sweep supplies endpoint and mechanism sensitivity
(Appendix~\ref{app:mech}).

\begingroup
\setlength{\intextsep}{4pt}
\begin{table}[b]
\centering
\small
\caption{All four tasks: ten cells and $240$ attempted
forks each; Forks counts those retained by the divergence rule, from which
Hopper is exempt (its floor-level return would make the rule ill-defined). Returns are
task-specific, never pooled; we average $\dR$ within cells and report checkpoint-cluster bootstrap
intervals. Drift columns: \emph{toward} decays the perturbation back toward the pretraining
physics; \emph{away} ramps it away from them. CartPole, Walker and
Cheetah use the corrected pre-fork sweep; Hopper ran the legacy saturating drift schedule (see
text) as the registered negative control.}
\label{tab:multitask}
\begin{tabular}{lrrrrr}
\toprule
 & & & & \multicolumn{2}{c}{mean $\dR$ by drift} \\
\cmidrule(lr){5-6}
Task & Cells & Forks (retained) & mean $\dR$ [95\% CI] & toward & away \\
\midrule
\texttt{cartpole-swingup} & 10 & 228 & $-113.4$ $[-131.2, -90.1]$ & $-152.6$ & $-74.1$ \\
\texttt{walker-walk}  & 10 & 238 & $-82.1$ $[-100.2, -60.9]$ & $-78.9$ & $-85.3$ \\
\texttt{cheetah-run}  & 10 & 227 & $-3.9$ $[-17.5, +13.0]$  & $-26.8$ & $+19.0$ \\
\texttt{hopper-hop}   & 10 & 240 & $+1.0$ $[-0.3, +2.4]$    & $+0.2$  & $+1.8$ \\
\bottomrule
\end{tabular}
\end{table}
\endgroup

\begingroup
\setlength{\intextsep}{4pt}
\begin{figure}[b]
\centering
\includegraphics[width=0.88\linewidth]{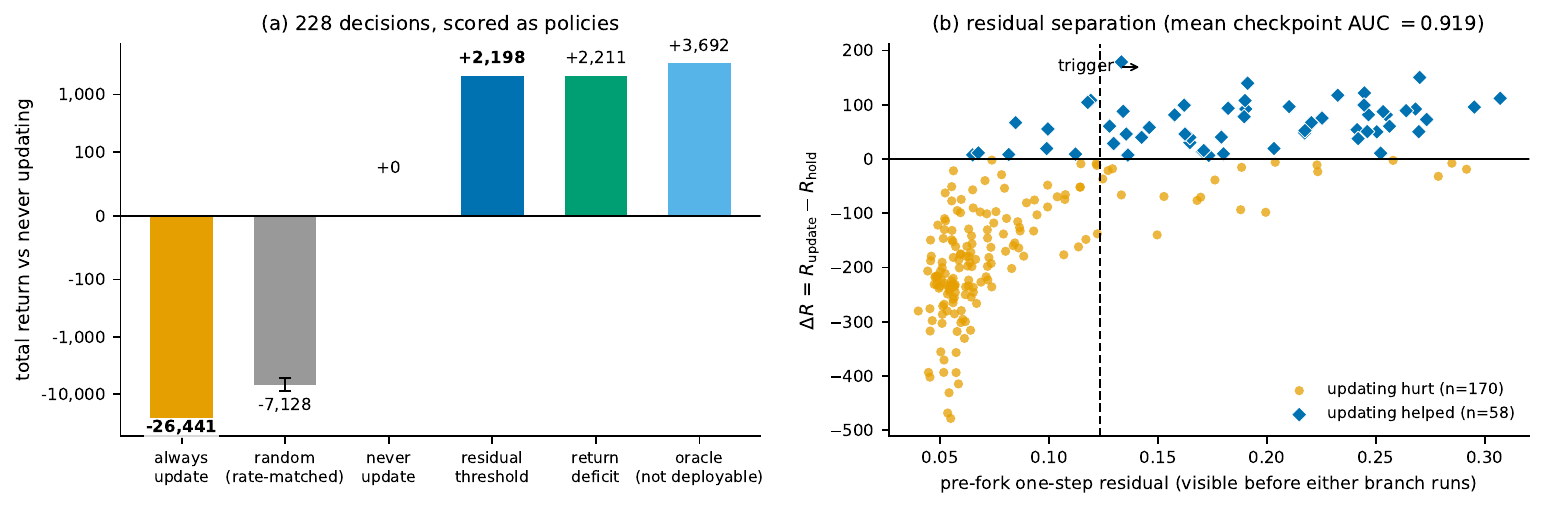}
\caption{\textbf{Update rules as policies on the corrected CartPole sweep.} \textbf{(a)} Total
return over $228$ non-diverged forks at $N=200$, relative to never updating (zero line), on a
symmetric-log axis (linear within $\pm100$); the
rate-matched random rule updates $70$ times and shows its $95\%$ permutation interval. \textbf{(b)}
Each point is one fork; points right of the dashed median leave-one-checkpoint-out threshold
trigger an update. The signal enriches helpful updates but does not separate them perfectly.}
\label{fig:policy}
\end{figure}
\endgroup

\subsection{Independent stream seeds preserve the task pattern}
\label{sec:seedrep}

Two additional stream seeds per checkpoint and drift direction add $60$ cells ($30$ per task
combined with the original ledgers). The sign pattern is unchanged: CartPole and Walker remain
negative in every cell, Cheetah unresolved (Table~\ref{tab:seedrep}). The replication supports
only the claim that those negative effects are not artifacts of a single stream realization; it
does not turn three task families into nine independent tasks.

\begingroup
\setlength{\intextsep}{4pt}
\begin{table}[b]
\centering
\small
\caption{Original corrected ledgers and independent stream-seed replication. Intervals are
checkpoint-clustered. The replication adds stream variation within each task; it does not change the
task-level uncertainty unit.}
\label{tab:seedrep}
\begin{tabular}{lcc}
\toprule
Task & original $10$ cells & combined $30$ cells \\
\midrule
CartPole & $-113.4$ $[-131.2,-90.1]$ & $-113.0$ $[-125.9,-97.4]$ \\
Walker   & $-82.1$ $[-100.2,-60.9]$  & $-81.1$ $[-98.4,-62.2]$ \\
Cheetah  & $-3.9$ $[-17.5,+13.0]$   & $-3.3$ $[-16.9,+10.8]$ \\
\bottomrule
\end{tabular}
\end{table}
\endgroup

\subsection{Why the update's value varies: a registered mechanism, falsified}
\label{sec:mechanism}

The CartPole result is not uniform across decision points. In that dose sweep,
regressing $R_{\textsc{update}}$ on $R_{\textsc{hold}}$ across $1048$ retained forks gives a slope
of $0.27$, not $1$: the updated branch retains a quarter of the variation in frozen performance, with
standard deviation $76$ versus $150$. The fitted line crosses the identity at
$R_{\textsc{hold}}=397$. Binned mean $\dR$ is $+55$ below $350$, $+7$ in $[350,450)$, then $-73$,
$-149$, and $-240$ as the frozen model improves (Figure~\ref{fig:attractor}), so most observed
decisions lie in the region where this update is harmful.

\begingroup
\setlength{\intextsep}{4pt}
\begin{figure}[b]
\centering
\includegraphics[width=0.64\linewidth]{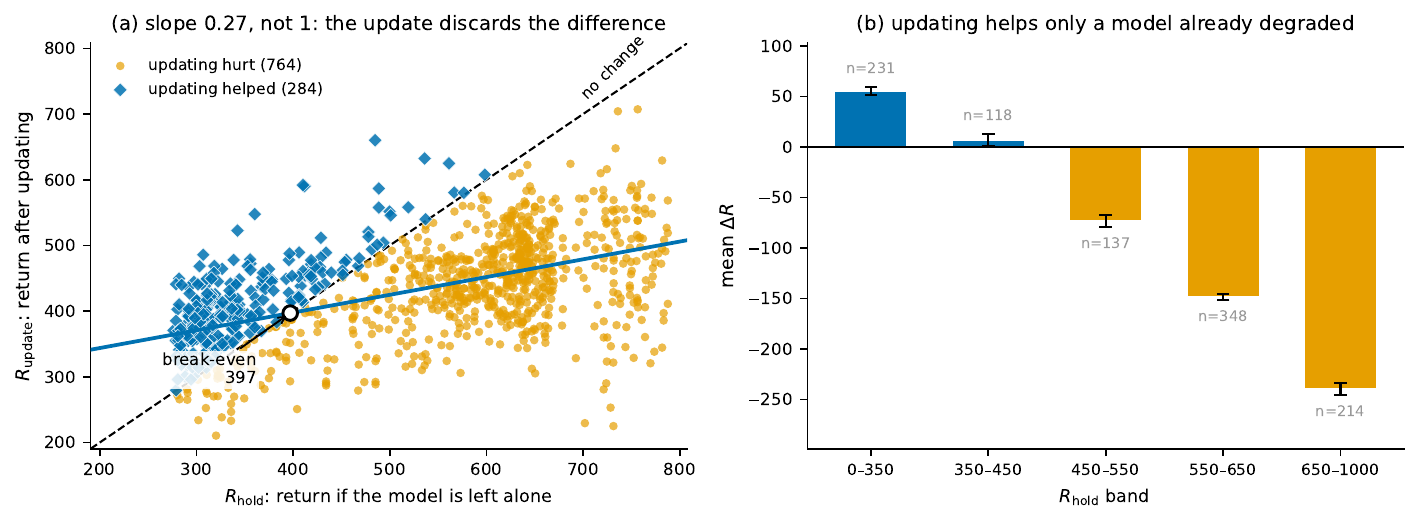}
\caption{\textbf{Update value versus frozen-branch return in the legacy CartPole dose sweep (exploratory; the registered transfer prediction built on this fit was falsified).}
\textbf{(a)} Each point is a retained fork across doses and checkpoints ($1048$ of $1088$); the fit
has slope $0.27$ and crosses the identity at $R_{\textsc{hold}}=397$. \textbf{(b)} Binned means
show the sign change: updating helps when the model is already degraded, costs more
as the frozen branch improves. Error bars are standard errors.}
\label{fig:attractor}
\end{figure}
\endgroup

Before running either additional task, we registered the following prediction: because Walker
($297.2$) and Cheetah ($278$) converge below CartPole's crossing at $397$, both should show $\dR$
between $+70$ and $+110$, with a pre-registered falsification threshold of $\dR < -50$ on either
task. The corrected sweep gives Walker $-82.1$, violating that criterion, and Cheetah
$-3.9$. Under the registered criterion the attractor account is therefore
refuted as a transferable mechanism, and everything in this subsection past that verdict is
exploratory. The per-task crossing analysis that follows is \emph{post hoc}:
fitting each task separately explains why the prediction failed. The crossing is a property of the
\emph{task}, not a transferable constant, but this analysis was not registered.

We also cannot claim the contraction generalises. The per-task legacy slopes, fit before the
divergence exclusion, are $0.28$ and $0.34$, close to CartPole's $0.27$, but their cell-clustered
intervals are $[-0.16, 0.76]$ and $[-0.43, 0.78]$. More decisively, a contraction toward a point
must \emph{shrink} the spread of outcomes, whereas the variance ratio is $3.14$ $[2.05,5.14]$ on
Walker, evidence of expansion, and $1.35$ $[0.71,3.97]$ on Cheetah. The slopes agree because
CartPole's competence range is much wider than Walker's, not because a common operator is at
work.

Across tasks, the crossing itself does not transfer. Walker is negative under both drift directions,
whereas Cheetah changes sign with the direction (Table~\ref{tab:multitask}), so the same update has
task- and drift-dependent value even with the
fork protocol held fixed. Measurement error and continual-learning pathologies such as loss of plasticity and primacy bias
are two alternative explanations that we test~\citep{dohare2024plasticity,nikishin2022primacy}; neither explains the
observed task- and direction-dependent pattern (Appendix~\ref{app:slope}).

The effect survives two objections to a single-cell result. Across that sweep's
$16\times$ dose range over five independently pretrained checkpoints, mean $\dR$ is negative
at every dose, and averaging within checkpoint leaves all five doses excluding zero (at $N=400$,
where four of the five checkpoints have cells, CI $[-159.0, -17.0]$, minimum detectable effect
$89.8$; Appendix~\ref{app:slope}). The corrected multi-task sweep shows the sign is not an
artifact of that dose grid, with Cheetah the boundary of the claim.

\subsection{No fixed-rate schedule beats never updating on CartPole}
\label{sec:schedules}

A rule blind to task, drift direction and current competence, the quantities the sections above
show $\dR$ tracks, cannot distinguish helpful from harmful updates, and neither family in use
reads them. That dose sweep is an analogue of DreamerV3's replay-ratio
knob~\citep{hafner2023dreamerv3}: at $N$ gradient steps every $250$ control steps, the effective
ratio $N/250$ runs from $0.10$ to $1.60$. The
analogy is limited: each arm sums independent one-shot fork contrasts on CartPole---a single
$N$-step update of the frozen reference-stream model on its most recent $W$ transitions---not a
compounded replay schedule with full-buffer replay, and the arms are unequally sized ($159$ forks
at the top of the range against $216$ at the bottom, with legacy and corrected CartPole ledgers
mixed at $N=200$), so the family totals bound the range rather than rank it. Every setting of the
ratio loses return against never updating, but per fork the loss runs from $-82.7$ to $-114.1$
with no monotone trend in $N$, and the decline in loss per gradient step across the range is the
denominator growing rather than the update improving. No arm compounds, and none applies the
periodic resets with which the replay-ratio literature buys higher ratios~\citep{dro2023breaking}. The corrected
CartPole sweep gives the matched policy totals in Figure~\ref{fig:policy}, at $200$ gradient
steps per update: both pre-fork signals recover positive utility, but neither reaches the oracle and this does not establish a
universal trigger; of the registered families, only residual and random appear here
(Section~\ref{sec:trigger} accounts for those absent from the corrected comparison). Periodic retraining is likewise monotone toward zero in the legacy CartPole schedule
baselines. All schedule evidence in this subsection is CartPole-only, from per-decision one-shot
contrasts.
This is a specification failure rather than a tuning failure: state-conditioned timing is useful
on corrected CartPole, but the cross-task results show its learned ordering is not automatically
portable. Learned replay scheduling can beat fixed schedules in continual
learning~\citep{klasson2023replaysched}; the ledger adds the realized per-decision value.

\subsection{Trigger rankings do not demonstrably transfer across tasks}
\label{sec:trigger}

The corrected sweep makes the comparison possible without leakage: we rank the pre-fork residual
and the deficit in mean return from the preceding deployment segment under the same
leave-one-checkpoint-out calibration and $\dR$ labels. The residual trigger was
registered; the return-deficit feature was added after registration and is exploratory. The
registered raw-KL signal, scored on the same ledgers and retention rule, tracks the residual
(AUC $0.917/0.502/0.674$ across the three tasks). The
registered drift oracle, a simulator-only ceiling, ranks better than either deployable signal
and earns less: AUC $0.937/0.409/0.701$ against policy totals $+1596/-1948/+800$, which is the
ranking-is-not-utility gap in one row. As a
deviation from the registration, four of the eight registered families are missing from this
comparison: the periodic schedule ran only on quarantined legacy
ledgers, and running variance, percentile, and held-out validation loss were never run
(Appendix~\ref{app:dev}). On CartPole the two policy signals we run are effectively
tied: AUC $0.919$ $[0.869,0.966]$ versus $0.915$ $[0.870,0.959]$, with policy totals $+2198$
and $+2211$ on $228$ retained forks against an oracle of $+3692$ (Appendix~\ref{app:tables}). A
rate-matched random residual rule earns $-7128$ $[-8949,-5303]$ ($20000$ cell-stratified
permutations), so the CartPole totals are not a generic consequence of
triggering; this control was pre-specified and run on all three tasks. Restricting the same
control to contiguous time blocks preserves the verdict wherever the restriction leaves the null
any freedom: the null total is $-523$ at two blocks and $+1599$ at four, both $p<0.001$ against
the observed $+2198$, while at six blocks each block holds three to four forks and the null
converges on the observed value by construction. The CartPole total is
also direction-dependent: the residual rule earns $+2033$ \emph{away}
versus $+165$ \emph{toward}, and the return-deficit rule $+1856$ versus $+355$,
so the positive aggregate is not a claim of equal benefit across drift directions.

\begingroup
\setlength{\intextsep}{4pt}
\begin{table}[b]
\centering
\footnotesize
\setlength{\tabcolsep}{3pt}
\caption{Pre-fork trigger comparison across the three corrected task sweeps, ten cells each. Both
signals are recorded before either branch runs and use the same leave-one-checkpoint-out threshold
rule; each random control is matched to its own policy's firing pattern. AUCs are
checkpoint-clustered; policy totals are task-specific and never pooled. Walker's AUC brackets are
min--max ranges over the only two checkpoint clusters containing a positive fork, not bootstrap
intervals (Section~\ref{sec:trigger}).}
\label{tab:trigger-rivals}
\begin{tabular}{lrrrrrr}
\toprule
Task & forks & residual AUC & return AUC & LOO $\dR$ (res. / ret.) & oracle & random control (res. / ret.) \\
\midrule
\texttt{cartpole} & $228$ & $.919$ $[.869,.966]$ & $.915$ $[.870,.959]$ & $+2198 / +2211$ & $+3692$ & $-7128 / -6889$ \\
\texttt{walker}   & $238$ & $.447$ $[.234,.659]$ & $.540$ $[.506,.574]$ & $-1010 / -681$ & $+21$ & $-945 / -618$ \\
\texttt{cheetah}  & $227$ & $.680$ $[.524,.812]$ & $.566$ $[.428,.670]$ & $+925 / +1310$ & $+3697$ & $-74 / +201$ \\
\bottomrule
\end{tabular}
\end{table}
\endgroup

The ordering is not universal, and the control does not pass everywhere. On Walker only five of
$238$ retained forks have $\dR>0$, so its AUCs are barely defined: three of the five checkpoint
clusters contain no positive fork, and the Walker brackets in Table~\ref{tab:trigger-rivals} are
min--max over the two that remain, one of them resting on a single positive fork; we draw no
ranking conclusion from them. Both learned
policies stay harmful ($-1010$ and $-681$) and each lies inside its own rate-matched control
($-945$ $[-1151,-750]$ and $-618$ $[-764,-475]$), so both fail there the check CartPole passes;
Walker's oracle is $+21$, so no rule has headroom. On Cheetah the two signals swap ($0.680$
versus $0.566$, totals $+925$ and $+1310$ against an oracle of $+3697$) and both controls are
cleared, but conditionally: with the divergence rule off the totals fall to $+16$ and $+100$ and
clear neither ($p=0.41$, $p=0.27$), because Cheetah's $13$ excluded forks sit at the $0.8$ quantile
of the retained residual distribution, where the trigger fires. CartPole's $12$ sit at $0.2$ and
its totals are unchanged, so CartPole is the only task that passes this control unconditionally. The data
support a protocol for testing triggers and a single-task demonstration that a pre-fork signal can
select less damaging moments, not a universal rule and not a cross-task reordering.

\paragraph{A ranking score is not policy utility.}
\label{sec:ranking-policy}
The AUC asks only for that ordering within a deployment stream; a policy must also turn the
ordering into a decision rule and pay the value of every selected update, so each threshold is
fit on four checkpoints and scored on the fifth.

An ordering can also be a proxy for something else. On CartPole the residual's correlation with
$\dR$ falls from $+0.756$ to $-0.125$ once $R_{\textsc{hold}}$ is held fixed, and both pre-fork
signals rank almost inversely with it (Spearman $-0.914$ and $-0.956$), so what they order there
is competence; on Cheetah the residual survives that conditioning ($+0.362$ raw, $+0.269$
partial). A
ranking diagnostic can establish that a signal is useful for a particular deployment stream; it
cannot by itself establish that the resulting policy is useful, portable, or optimal.

We also do not pool these totals: available oracle value is roughly $+3{,}700$ on CartPole and
Cheetah but $+21$ on Walker, so an aggregate would follow wherever the simulator left headroom.

\section{Discussion}
\label{sec:limitations}
\label{sec:discussion}

The fork ledger supplies the check that aggregate-return comparisons omit: the realized value of
applying one update at one decision point. In this study, the sign of update value varies with task
and drift direction, quantities a fixed replay ratio does not read. The CartPole trigger
comparison shows that a signal can select useful updates within a task; Walker and Cheetah show why
that ordering should be tested rather than assumed to transfer. By design the instrument never
runs a continually adapting model: the stream continues from the \textsc{hold} branch, so each
$\dR$ prices one update against a never-updated reference. A committing variant, in which fired
updates persist, is the natural successor instrument.

\paragraph{What the paired ledger isolates.} On CartPole the rate-matched random rule and the
residual trigger spend the same $14{,}000$ gradient steps, yet one loses $-7{,}128$ return units
and the other earns $+2{,}198$. With matched counterfactual branches, that difference measures
whether the rule chose better moments for the same intervention, an inference a trajectory-level
aggregate conceals.

\paragraph{Scope and disclosures.} The ledger needs restorable simulator state, a scalar
downstream outcome, and one fixed update mechanism. Proprioceptive control supplies all three
cheaply; interactive video world models supply the first two and lack only an agreed downstream
outcome. It is an evaluation instrument, not a deployable trigger policy; the registered drift oracle is scored here but is simulator-only,
since a deployment cannot read its own drift level. Three tasks carry return dynamics;
the task-level result is therefore that the sign varies across the studied tasks and drift directions,
not a population estimate. The legacy-schedule Hopper control is exempt from the divergence rule
because its floor-level return would make the rule ill-defined. The registered
arms with no committed ledger are enumerated in Appendix~\ref{app:dev}. The parallel-episode cost model was
calibrated only to $30$ episodes, below the $60$ per fork our cells run.

\typeout{CWMBODYENDSONPAGE=\thepage}
\clearpage
\bibliographystyle{plainnat}
\bibliography{refs}

@article{hafner2023dreamerv3,
  title={Mastering diverse control tasks through world models},
  author={Hafner, Danijar and Pasukonis, Jurgis and Ba, Jimmy and Lillicrap, Timothy},
  journal={Nature},
  volume={640},
  pages={647--653},
  year={2025},
  doi={10.1038/s41586-025-08744-2},
  url={https://doi.org/10.1038/s41586-025-08744-2}
}

@inproceedings{hansen2024tdmpc2,
  title={{TD-MPC2}: Scalable, Robust World Models for Continuous Control},
  author={Hansen, Nicklas and Su, Hao and Wang, Xiaolong},
  booktitle={International Conference on Learning Representations (ICLR)},
  year={2024},
  url={https://arxiv.org/abs/2310.16828}
}

@article{tassa2018dmc,
  title={{DeepMind} Control Suite},
  author={Tassa, Yuval and Doron, Yotam and Muldal, Alistair and Erez, Tom and Li, Yazhe and de Las Casas, Diego and Budden, David and Abdolmaleki, Abbas and Merel, Josh and Lefrancq, Andrew and Lillicrap, Timothy and Riedmiller, Martin},
  journal={arXiv preprint arXiv:1801.00690},
  year={2018},
  url={https://arxiv.org/abs/1801.00690}
}

@inproceedings{todorov2012mujoco,
  title={{MuJoCo}: A physics engine for model-based control},
  author={Todorov, Emanuel and Erez, Tom and Tassa, Yuval},
  booktitle={IEEE/RSJ International Conference on Intelligent Robots and Systems (IROS)},
  pages={5026--5033},
  year={2012},
  organization={IEEE},
  doi={10.1109/IROS.2012.6386109},
  url={https://doi.org/10.1109/IROS.2012.6386109}
}

@inproceedings{pathak2019disagreement,
  title={Self-Supervised Exploration via Disagreement},
  author={Pathak, Deepak and Gandhi, Dhiraj and Gupta, Abhinav},
  booktitle={Proceedings of the 36th International Conference on Machine Learning},
  series={Proceedings of Machine Learning Research},
  volume={97},
  pages={5062--5071},
  publisher={PMLR},
  year={2019},
  url={https://proceedings.mlr.press/v97/pathak19a.html}
}

@inproceedings{mavorparker2022noisytv,
  title={How to Stay Curious while avoiding Noisy {TV}s using Aleatoric Uncertainty Estimation},
  author={Mavor-Parker, Augustine N. and Young, Kimberly A. and Barry, Caswell and Griffin, Lewis D.},
  booktitle={Proceedings of the 39th International Conference on Machine Learning},
  series={Proceedings of Machine Learning Research},
  volume={162},
  pages={15220--15240},
  publisher={PMLR},
  year={2022},
  url={https://proceedings.mlr.press/v162/mavor-parker22a.html}
}

@inproceedings{jarrett2023hindsight,
  title={Curiosity in Hindsight: Intrinsic Exploration in Stochastic Environments},
  author={Jarrett, Daniel and Tallec, Corentin and Altch{\'e}, Florent and Mesnard, Thomas and Munos, R{\'e}mi and Valko, Michal},
  booktitle={Proceedings of the 40th International Conference on Machine Learning},
  series={Proceedings of Machine Learning Research},
  volume={202},
  pages={14780--14816},
  publisher={PMLR},
  year={2023},
  url={https://proceedings.mlr.press/v202/jarrett23a.html}
}

@inproceedings{nalisnick2019know,
  title={Do Deep Generative Models Know What They Don't Know?},
  author={Nalisnick, Eric and Matsukawa, Akihiro and Teh, Yee Whye and Gorur, Dilan and Lakshminarayanan, Balaji},
  booktitle={International Conference on Learning Representations (ICLR)},
  year={2019},
  url={https://arxiv.org/abs/1810.09136}
}

@inproceedings{caterini2021entropic,
  title={Entropic Issues in Likelihood-Based {OOD} Detection},
  author={Caterini, Anthony L. and Loaiza-Ganem, Gabriel},
  booktitle={Proceedings on ``I (Still) Can't Believe It's Not Better!'' at NeurIPS 2021 Workshops},
  series={Proceedings of Machine Learning Research},
  volume={163},
  pages={21--26},
  publisher={PMLR},
  year={2022},
  url={https://proceedings.mlr.press/v163/caterini22a.html}
}

@inproceedings{zhang2021understanding,
  title={Understanding Failures in Out-of-Distribution Detection with Deep Generative Models},
  author={Zhang, Lily H. and Goldstein, Mark and Ranganath, Rajesh},
  booktitle={Proceedings of the 38th International Conference on Machine Learning},
  series={Proceedings of Machine Learning Research},
  volume={139},
  pages={12427--12436},
  publisher={PMLR},
  year={2021},
  url={https://proceedings.mlr.press/v139/zhang21g.html}
}

@inproceedings{lambert2020objective,
  title={Objective Mismatch in Model-based Reinforcement Learning},
  author={Lambert, Nathan and Amos, Brandon and Yadan, Omry and Calandra, Roberto},
  booktitle={Proceedings of the 2nd Conference on Learning for Dynamics and Control},
  series={Proceedings of Machine Learning Research},
  volume={120},
  pages={761--770},
  publisher={PMLR},
  year={2020},
  url={https://proceedings.mlr.press/v120/lambert20a.html}
}

@inproceedings{dorka2023dutd,
  title={Dynamic Update-to-Data Ratio: Minimizing World Model Overfitting},
  author={Dorka, Nicolai and Welschehold, Tim and Burgard, Wolfram},
  booktitle={International Conference on Learning Representations (ICLR)},
  year={2023},
  note={arXiv:2303.10144},
  url={https://arxiv.org/abs/2303.10144}
}

@inproceedings{ji2022cmlo,
  title={When to Update Your Model: Constrained Model-based Reinforcement Learning},
  author={Ji, Tianying and Luo, Yu and Sun, Fuchun and Jing, Mingxuan and He, Fengxiang and Huang, Wenbing},
  booktitle={Advances in Neural Information Processing Systems (NeurIPS)},
  year={2022},
  note={arXiv:2210.08349},
  url={https://arxiv.org/abs/2210.08349}
}

@inproceedings{vovk2021retrain,
  title={Retrain or not retrain: Conformal test martingales for change-point detection},
  author={Vovk, Vladimir and Petej, Ivan and Nouretdinov, Ilia and Ahlberg, Ernst and Carlsson, Lars and Gammerman, Alex},
  booktitle={Conformal and Probabilistic Prediction and Applications (COPA), PMLR 152},
  pages={191--210},
  year={2021},
  url={https://proceedings.mlr.press/v152/vovk21b/vovk21b.pdf}
}

@inproceedings{nasvytis2024rethinking,
  title={Rethinking Out-of-Distribution Detection for Reinforcement Learning: Advancing Methods for Evaluation and Detection},
  author={Nasvytis, Linas and Sandbrink, Kai and Foerster, Jakob and Franzmeyer, Tim and Schroeder de Witt, Christian},
  booktitle={International Conference on Autonomous Agents and Multiagent Systems (AAMAS)},
  year={2024},
  note={arXiv:2404.07099},
  url={https://arxiv.org/abs/2404.07099}
}

@inproceedings{zollicoffer2025novelty,
  title={Novelty Detection in Reinforcement Learning with World Models},
  author={Zollicoffer, Geigh and Eaton, Kenneth and Balloch, Jonathan C. and Kim, Julia and Zhou, Wei and Wright, Robert and Riedl, Mark},
  booktitle={Proceedings of the 42nd International Conference on Machine Learning (ICML)},
  series={Proceedings of Machine Learning Research},
  volume={267},
  pages={80740--80758},
  publisher={PMLR},
  year={2025},
  url={https://proceedings.mlr.press/v267/zollicoffer25a.html}
}

@inproceedings{feng2023fowm,
  title={Finetuning Offline World Models in the Real World},
  author={Feng, Yunhai and Hansen, Nicklas and Xiong, Ziyan and Rajagopalan, Chandramouli and Wang, Xiaolong},
  booktitle={Conference on Robot Learning (CoRL)},
  year={2023},
  note={arXiv:2310.16029},
  url={https://arxiv.org/abs/2310.16029}
}

@article{rubinstein1999cem,
  title={The Cross-Entropy Method for Combinatorial and Continuous Optimization},
  author={Rubinstein, Reuven},
  journal={Methodology and Computing in Applied Probability},
  volume={1},
  pages={127--190},
  year={1999},
  doi={10.1023/A:1010091220143},
  url={https://link.springer.com/article/10.1023/A:1010091220143}
}

@article{glasserman1992crn,
  title={Some Guidelines and Guarantees for Common Random Numbers},
  author={Glasserman, Paul and Yao, David D.},
  journal={Management Science}, volume={38}, number={6}, pages={884--908}, year={1992},
  doi={10.1287/mnsc.38.6.884},
  url={https://doi.org/10.1287/mnsc.38.6.884}
}

@article{predictability2026control,
  title={A Control Theory of Predictability in Latent World Models},
  author={You, Hanzhe and Zhang, Yonggang and Ran, Maohao and Yang, Zhiqin and Zhang, Zhenyuan and Xue, Wei and Song, Jun and Tian, Xinmei and Guo, Yike},
  year={2026}, journal={arXiv preprint}, note={arXiv:2607.10362},
  url={https://arxiv.org/abs/2607.10362}
}

@article{cameron2008clusterboot,
  title={Bootstrap-Based Improvements for Inference with Clustered Errors},
  author={Cameron, A. Colin and Gelbach, Jonah B. and Miller, Douglas L.},
  journal={Review of Economics and Statistics}, volume={90}, number={3},
  pages={414--427}, year={2008},
  doi={10.1162/rest.90.3.414},
  url={https://doi.org/10.1162/rest.90.3.414}
}

@article{anderson2003permutation,
  title={Permutation tests for multi-factorial analysis of variance},
  author={Anderson, Marti J. and ter Braak, Cajo J. F.},
  journal={Journal of Statistical Computation and Simulation},
  volume={73}, number={2}, pages={85--113}, year={2003},
  doi={10.1080/00949650215733},
  url={https://doi.org/10.1080/00949650215733}
}

@article{dohare2024plasticity,
  title   = {Loss of plasticity in deep continual learning},
  author  = {Dohare, Shibhansh and Hernandez-Garcia, J. Fernando and Lan, Qingfeng and
             Rahman, Parash and Mahmood, A. Rupam and Sutton, Richard S.},
  journal = {Nature},
  volume  = {632},
  pages   = {768--774},
  year    = {2024},
  doi     = {10.1038/s41586-024-07711-7},
  url     = {https://doi.org/10.1038/s41586-024-07711-7}
}

@inproceedings{nikishin2022primacy,
  title     = {The Primacy Bias in Deep Reinforcement Learning},
  author    = {Nikishin, Evgenii and Schwarzer, Max and D'Oro, Pierluca and
               Bacon, Pierre-Luc and Courville, Aaron},
  booktitle = {Proceedings of the 39th International Conference on Machine Learning},
  series    = {Proceedings of Machine Learning Research},
  volume    = {162},
  pages     = {16828--16847},
  publisher = {PMLR},
  year      = {2022},
  url       = {https://proceedings.mlr.press/v162/nikishin22a.html}
}

@inproceedings{oberst2019counterfactual,
  title     = {Counterfactual Off-Policy Evaluation with {G}umbel-Max Structural Causal Models},
  author    = {Oberst, Michael and Sontag, David},
  booktitle = {International Conference on Machine Learning (ICML)},
  series    = {PMLR},
  volume    = {97},
  pages     = {4881--4890},
  year      = {2019},
  note      = {arXiv:1905.05824},
  url       = {https://proceedings.mlr.press/v97/oberst19a.html}
}

@article{domberg2026selfadapting,
  title={Self-adapting Robotic Agents through Online Continual Reinforcement Learning with World Model Feedback},
  author={Domberg, Fabian and Schildbach, Georg},
  journal={arXiv preprint arXiv:2603.04029},
  year={2026},
  url={https://arxiv.org/abs/2603.04029}
}

@article{twinrollouts2026,
  title={Twin Rollouts: Noise-Coupled Counterfactual Branching in Interactive Video World Models},
  author={Ma, Yu and Shi, Hongli and Xu, Xinran},
  journal={arXiv preprint arXiv:2608.08982},
  year={2026},
  url={https://arxiv.org/abs/2608.08982}
}

@article{coco2026,
  title={Overcoming Statistical Bias in Action-Controllable World Models},
  author={Shi, Yuhong and Chu, Zhenhao and Wei, Jie and Hao, Jun and Liu, Jianyi and Fu, Jingwen},
  journal={arXiv preprint arXiv:2608.04653},
  year={2026},
  url={https://arxiv.org/abs/2608.04653}
}

@inproceedings{buesing2019cfgps,
  title={Woulda, Coulda, Shoulda: Counterfactually-Guided Policy Search},
  author={Buesing, Lars and Weber, Th{\'e}ophane and Zwols, Yori and Racani{\`e}re, S{\'e}bastien and Guez, Arthur and Lespiau, Jean-Baptiste and Heess, Nicolas},
  booktitle={International Conference on Learning Representations (ICLR)},
  year={2019},
  url={https://arxiv.org/abs/1811.06272}
}

@inproceedings{dro2023breaking,
  title     = {Sample-Efficient Reinforcement Learning by Breaking the Replay Ratio Barrier},
  author    = {D'Oro, Pierluca and Schwarzer, Max and Nikishin, Evgenii and
               Bacon, Pierre-Luc and Bellemare, Marc G. and Courville, Aaron},
  booktitle = {International Conference on Learning Representations (ICLR)},
  year      = {2023},
  url       = {https://iclr.cc/virtual/2023/oral/12655}
}

@article{regol2025retrain,
  title   = {When to retrain a machine learning model},
  author  = {Regol, Florence and Schwinn, Leo and Sprague, Kyle and Coates, Mark and
             Markovich, Thomas},
  journal = {arXiv preprint arXiv:2505.14903},
  year    = {2025},
  url     = {https://arxiv.org/abs/2505.14903}
}

@article{steland2026online,
  title   = {Online Detection of Changes in Moment-Based Projections: When to Retrain
             Deep Learners or Update Portfolios?},
  author  = {Steland, Ansgar},
  journal = {Journal of Machine Learning Research},
  volume  = {27},
  pages   = {1--50},
  year    = {2026},
  url     = {https://www.jmlr.org/papers/volume27/23-0274/23-0274.pdf}
}

@inproceedings{liu2025continualwm,
  title     = {Continual Reinforcement Learning by Planning with Online World Models},
  author    = {Liu, Zichen and Fu, Guoji and Du, Chao and Lee, Wee Sun and Lin, Min},
  booktitle = {Proceedings of the 42nd International Conference on Machine Learning},
  series    = {Proceedings of Machine Learning Research},
  volume    = {267},
  pages     = {38397--38423},
  publisher = {PMLR},
  year      = {2025},
  note      = {arXiv:2507.09177},
  url       = {https://proceedings.mlr.press/v267/liu25p.html}
}

@article{klasson2023replaysched,
  title   = {Learn the Time to Learn: Replay Scheduling in Continual Learning},
  author  = {Klasson, Marcus and Kjellstr{\"o}m, Hedvig and Zhang, Cheng},
  journal = {Transactions on Machine Learning Research},
  year    = {2023},
  note    = {arXiv:2209.08660},
  url     = {https://arxiv.org/abs/2209.08660}
}

@inproceedings{liebman2018stitch,
  title={A Stitch in Time - Autonomous Model Management via Reinforcement Learning},
  author={Liebman, Elad and Zavesky, Eric and Stone, Peter},
  booktitle={Proceedings of the 17th International Conference on Autonomous Agents and MultiAgent Systems (AAMAS)},
  pages={990--998},
  year={2018},
  url={https://aamas.csc.liv.ac.uk/Proceedings/aamas2018/pdfs/p990.pdf}
}

\appendix

\section{Protocol and robustness notes}
\label{app:notes}

\subsection{Primary protocol}\label{app:prereg}
The corrected primary arm uses one fixed update response: $N=200$ gradient steps on the most recent
$W=10{,}000$ transitions, evaluated on $m=30$ matched episodes per branch at $\sa=0$. The corrected
grid per task is five pretrained checkpoints $\times$ two drift directions ($10$ cells and $24$
attempted forks per cell) under a $20{,}000$-step drift period. The model is pretrained only on
clean dynamics; every paired continuation pins the initial state, injected-action-noise stream,
and recurrent-model sampler. No trigger, threshold, or update setting is chosen from the ledger it
later scores. The accompanying code artifact contains the harness, the analysis scripts behind the
corrected results, and the corrected fork ledgers, whose JSON records the executed configuration of
every cell.

\subsection{Deviations from the registered design}\label{app:dev}
The registered evaluation-episode ramp ($30, 75, 120$ across the three $\sa$ cells) was not
executed: every committed corrected-arm ledger runs $m=30$ at $\sa=0$, so the stochasticity axis of
Section~\ref{sec:setting} is unexercised in the corrected arm (the quarantined legacy CartPole
ledgers shipped for sensitivity include $\sa\in\{0.3,0.6,1.2\}$ cells). Of the eight registered trigger
families (thresholded one-step error, running variance, percentile, raw KL, held-out validation
loss, periodic, random, and drift oracle), the corrected arm scores four---thresholded one-step
error, raw KL (as a ranking signal, Section~\ref{sec:trigger}), the drift oracle, and
rate-matched random triggering---alongside one unregistered addition, the deficit in
pre-fork deployment return, and the $\dR$ oracle ceiling. The periodic schedule has results only
on the quarantined legacy ledgers; running variance, percentile,
and held-out validation loss have no result anywhere. The registered causal running-quantile
threshold rule was not used: corrected thresholds are fit leave-one-checkpoint-out on the
pre-fork signal (Section~\ref{sec:design}). The registered primary endpoint is not reported
(Section~\ref{sec:design}). Hopper's negative control ran the legacy $2{,}000$-step
drift schedule rather than the corrected $20{,}000$-step schedule (Section~\ref{sec:cost}).

\subsection{Policy and trigger totals}\label{app:tables}
On the $228$ retained CartPole forks, always updating totals $-26{,}441$ return units relative to
never updating. Leave-one-checkpoint-out residual and return-deficit triggers total $+2{,}198$ and
$+2{,}211$, respectively; the unachievable $\dR$ oracle totals $+3{,}692$. Across tasks, residual
and return-deficit AUCs are nearly tied on CartPole ($.919$ and $.915$), are defined on only two of
five checkpoint clusters on Walker ($.447$ and $.540$, min--max ranges rather than intervals), and
overlap on Cheetah ($.680$ and $.566$). Returns are never pooled across tasks.

\subsection{Endpoint and mechanism robustness}\label{app:mech}
The corrected \emph{CartPole} endpoint remains negative at three update doses: mean cell-level
$\dR=-88.4$ ($[-102.0,-71.9]$) at $N=50$, $-113.4$ ($[-131.2,-90.1]$) at $N=200$, and $-110.1$
($[-132.2,-86.6]$) at $N=400$; the dose arm was not run on Walker or Cheetah, so dose insensitivity
is established only for CartPole. A separate $100$-cell update-mechanism sweep varying the window
to $W=2{,}000$ and $W=50{,}000$ changes the magnitude of the harm and, on \texttt{cheetah}
ascending, its sign, so the body's labels are conditional on the fixed mechanism. The independent
stream-seed replication preserves the negative CartPole and Walker effects while leaving Cheetah
unresolved.

\subsection{Measurement error, plasticity, and the dose interval}\label{app:slope}
Every interval in the body is a percentile bootstrap over five checkpoint clusters, where the
$2.5\%$ quantile sits near the smallest cluster mean and coverage is optimistic. The Student-$t$
intervals on the same five means are $[-146.7,-80.0]$ on CartPole, $[-111.7,-52.5]$ on Walker and
$[-28.4,+20.6]$ on Cheetah, so the two negative sign verdicts hold under either choice and
Cheetah stays unresolved under both. No multiplicity adjustment is applied across the signal
comparisons, which we read descriptively.
Splitting each fork's frozen episodes into independent halves
reproduces the identical noise structure with no update present and gives a null slope of $0.98$
($[0.98, 0.99]$) with a variance ratio of $1.00$, against $0.47$ and $0.44$ observed on the
same fork population (all $1821$ divergence-retained forks across the three tasks, pooled
only for this noise diagnostic), so measurement error does not produce contraction of the
kind in Section~\ref{sec:mechanism}. On the legacy CartPole sweep, holding
$R_{\textsc{hold}}$ fixed, the partial correlation of $\dR$ with fork index is $+0.09$, and
sixteen of seventeen comparable cells contain both signs of $\dR$, which monotone cumulative
plasticity loss cannot produce. The $N=400$ dose interval $[-159.0, -17.0]$ averages over the
four checkpoints with cells at that dose and uses $t_{.975,3}=3.182$; an earlier
``underpowered'' reading used $t_{.975,2}=4.303$ from a shifted degrees-of-freedom row.

\subsection{Pinned-physics bias and the forward-window arm}\label{app:bias}
In a toy staleness model the forward/pinned contrast ratio is $1.000$ for linear control loss,
rises to $1.98$--$10.8$ for convex loss as drift increases, and reverses under saturating losses
($0.791\to0.209$), so curvature, not drift rate alone, sets the direction of the pinning bias.
The completed forward-window arm bounds it empirically on CartPole: over ten matched cells,
forward minus pinned mean cell-level $\dR$ is $+3.9$ (checkpoint-bootstrap 95\% CI
$[-2.4,+10.2]$), and the leave-one-checkpoint-out policy totals remain close ($+2042$ versus
$+2198$ for the residual trigger, $+2210.6$ versus $+2211.4$ for the return deficit). The
CartPole headline endpoint is therefore not explained by pinning the evaluation physics; this
arm was not run on Walker or Cheetah, so it does not clear them, and the interval leaves small
sensitivity unresolved. The larger analyses---the continuous-endpoint derivation, common-random-number audits,
and the $\beta$ sensitivity---do not fit in this workshop paper; they qualify the scope of the
measurement but do not add a second headline result.

\end{document}